\documentclass{article}
\usepackage{times}
\usepackage{graphicx}
\usepackage{subfigure} 
\usepackage{natbib}
\usepackage{algorithm}
\usepackage{algorithmic}
\usepackage{hyperref}

\usepackage{amsmath, amsfonts}
\usepackage{bm} 

\newcommand{\xbacki}{\mathbf{x}_{\backslash i}}

\usepackage[accepted]{icml2016}
\icmltitlerunning{A New Method to Visualize Deep Neural Networks}

\begin{document} 

\twocolumn[
\icmltitle{A New Method to Visualize Deep Neural Networks}

\icmlauthor{Luisa M Zintgraf$^{1}$}{lmzintgraf@gmail.com}
\icmlauthor{\vspace{0.05cm} Taco S Cohen$^{1}$}{t.s.cohen@uva.nl}
\icmlauthor{\vspace{0.05cm} Max Welling$^{1, 2}$}{m.welling@uva.nl}
\icmladdress{
\vspace{0.1cm}$^1$Informatics Institute, University of Amsterdam \\ 
$^2$Canadian Institute for Advanced Research}

\icmlkeywords{deepvis, deep visualization, machine learning}
\vskip 0.3in ]

% ------------------- ABSTRACT ------------------------------------------------------------

\begin{abstract}
We present a method for visualizing the response of a deep neural network to a specific input. For image data for instance our method will highlight areas that provide evidence in favor of, and against choosing a certain class. The method overcomes several shortcomings of previous methods and provides great additional insight into the decision-making process of convolutional networks, which is important both to improve models and to accelerate the adoption of such methods in e.g. medicine. In experiments on ImageNet data, we illustrate how the method works and can be applied in different ways to understand deep neural nets.
\end{abstract}

% ------------------- INTRO ------------------------------------------------------------

\vspace{-0.5cm}
\section{Introduction} \label{introduction}

Deep convolutional neural networks (DCNNs) are classifiers tailored to the task of image recognition. In recent years, they have become increasingly powerful and deliver state-of-the-art performance on natural image classification tasks such as the ILSVRC competition \cite{ILSVRC15}. Achieving these results requires a well-chosen architecture for the network and fine-tuning its parameters correctly during training. Along with advances in computing power (efficient GPU implementations) and smarter training techniques, these networks have not only become better and feasible to train, but also much deeper and larger to achieve such high performances.

Consequently, these ever larger networks come at a price: it is very hard to comprehend how they operate, and what exactly it is that makes them so powerful, even if we understand the data well (e.g., images). To date, training a deep neural network (DNN) involves a lot of trial-and-error, until a satisfying set of parameters is found. The resulting networks resemble complex non-linear mathematical functions with millions of parameters, making it difficult to point out strengths or weaknesses. Not only is this unsatisfactory in itself, it also makes it much more difficult to improve the networks. 
A second argument for powerful visualization methods is the circumstance that the adoption of black-box methods such as deep neural networks in industry, government and healthcare is complicated by the fact that their responses are very difficult to understand. Imagine a physician using a DNN to diagnose a patient. S/he will most likely not trust an automated diagnosis unless s/he understands the reason behind a certain prediction (e.g. highlighted regions in the brain that differ from normal subjects) allowing him/her to verify the diagnosis and reason about it. 

Thus, methods for visualizing the decision-making process and inner workings of deep neural networks networks can be of great value for their qualitative assessment. Understanding them better will enable us to find new ways to guide training into the right direction and improve existing successful networks by detecting their weaknesses, as well as accelerating their adoption in society. 

This paper builds on a collection of new and intriguing methods for analyzing deep neural networks through visualization, which has emerged in the literature recently. We present a novel visualization method, exemplified for DCNNs, that finds and highlights the regions in image space that activate the nodes (hidden and output) in the neural network. Thus, it illustrates what the different parts of the network are looking for, given a specific input image. We show several examples of how the method can be deployed and used to understand the classifier's decision. 

\newpage

% ------------------- RELATED WORK---------------------------------------------------

\section{Related Work}

Understanding a DCNN by deep visualization can be approached from two perspectives, yielding different insights into how the network operates. 

First, we can try to find the \textit{notion of a class} (or unit) the network has, by finding an input image that maximizes the class (or unit) of interest. The resulting image gives us a sense of what excites the unit the most, i.e., what it is looking for, and is especially appealing when natural image priors are incorporated into the optimization procedure \cite{erhan2009visualizing, simonyan2013deep, yosinski2015understanding}.

The second option is to visualize how the network responds to a specific image, which will be the further subject of this paper.

\subsection{Instance-Specific Visualization} \label{sec:related:instance-specific}

Explaining how a DCNN makes decisions for a specific image can be visualized directly in image space by finding the spatial support of the prediction present in that image. 

Simonyan et al. \yrcite{simonyan2013deep} propose \textit{image-specific class saliency visualization} to rank the input features based on their influence on the assigned class $c$. To this end, they compute the partial derivative of the (pre-softmax) class score $S_c$ with respect to the input features $x_i$ (which are typically pixels), 
$ \displaystyle s_i = \partial S_{c} / \partial x_i , $
to estimate each feature's relevance. In words, this expresses how sensitive the classifier's prediction is to small changes of feature values in input space. The authors also show that there is a close connection to the visualization with the help of deconvolutional networks, proposed by Zeiler and Fergus \yrcite{zeiler2014visualizing}.

Zhou et al. \yrcite{zhou2014object} generate, for a specific input image, a simplified version of that image that still gets classified correctly, to visualize the regions most important for the classification. They do this by iteratively removing segments of the image and thus keeping as little visual information as possible.

The method we will present in this paper also visualizes the spatial support of a class (or unit) of the network directly in image space, for a given image. The idea is similar to an analysis Zeiler and Fergus \yrcite{zeiler2014visualizing} make, where they estimate the importance of input regions in the image by visualising the probability of the correct class as a function of a gray patch that is occluding parts of the image. In this paper, we will take a more rigorous approach at removing information from the image and observing how the network responds, based on a method by Robnik-{\v{S}}ikonja and Kononenko \yrcite{robnik2008explaining}. 

\subsection{Difference of Probabilities}

Robnik-{\v{S}}ikonja and Kononenko \yrcite{robnik2008explaining} propose a method for explaining predictions of probabilistic classifiers, given a \textit{specific} input instance. To express what role the input features took in the decision, each input feature is assigned a \textit{relevance} value for the given prediction (with respect to a class, e.g., the highest scoring one). This way, the method produces a relevance vector that is of the same size as the input, and which reflects the relative importance of all input features.

In order to measure how important a particular feature is for the decision, the authors look at how the prediction changes if this feature was unknown, i.e., the difference between $p(c|\mathbf{x})$ and $p(c|\xbacki)$ for a feature $x_i$ and a class $c$. The underlying idea is that if there is a large prediction difference, the feature must be important, whereas if there is little to no difference, the particular feature value has not contributed much to the class $c$.

While the concept is quite straightforward, evaluating the classifier's prediction when a feature is unknown is not. Only few classifiers can handle unknown values directly. Thus, to estimate the class probability $p(c|\xbacki)$ where feature $x_i$ is unknown, the authors propose a way to simulate the absence of a feature by approximately marginalizing it out. Marginalizing out a feature mathematically is given by
\begin{equation} \label{eq:marginal-exact}
p(c|\xbacki) = \sum\limits_{x_i} p(x_i|\xbacki) p(c|\xbacki, x_i ) \ ,
\end{equation}
and the authors choose to approximate this equation by
\begin{equation} \label{eq:marginal-approx}
p(c|\xbacki) \approx \sum\limits_{x_i} p(x_i) p(c|\xbacki, x_i) \ .
\end{equation}
The underlying assumption in $p(x_i|\xbacki)\approx p(x_i)$ is that feature $x_i$ is independent of the other features, $\xbacki$. In practice, the prior probability $p(x_i)$ is usually approximated by the empirical distribution for that feature.

Once the class probability $p(c|\xbacki)$ is estimated, it can be compared to $p(c|\mathbf{x})$. We will stick to an evaluation proposed by the authors referred to as \textit{weight of evidence}, given by
\begin{equation} \label{eq:weight-of-evidence}
\text{WE}_i(c|\mathbf{x})=\log_2\left(\text{odds}(c|\mathbf{x})\right)-\log_2\left(\text{odds}(c|\xbacki)\right) ,
\end{equation}
where
\begin{equation} \label{eq:log-odds}
\text{odds}(c|\mathbf{x}) = \frac{p(c|\mathbf{x})}{1-p(c|\mathbf{x})} \ .
\end{equation}
To avoid problems with zero-valued probabilities, they use the Laplace correction $\displaystyle p\leftarrow (pN+1)/(N+K)$, where $N$ is the number of training instances and $K$ is the number of classes.

\newpage
The resulting relevance vector has positive and negative entries. A positive value means that the corresponding feature has contributed \textit{towards} the class of interest. A negative value on the other hand means that the feature value was actually evidence \textit{against} the class (if it was unknown, the classifier would become more certain about the class under consideration because evidence against it was removed).

There are two main drawbacks to this method which we want to address in this paper. First, this is a \textit{univariate} approach: only one feature at a time is removed. We would expect that a neural network will not be so easily fooled and change its prediction if just one pixel value of a high-dimensional input was unknown, like a pixel in an image. Second, the approximation in equation (\ref{eq:marginal-approx}) is not very accurate. In this paper, we will show how this method can be improved when working with image data.

Furthermore, we will show how the method can not only be used to analyses the prediction outcome, but also for visualizing the role of hidden layers of deep neural networks.

% ------------------- APPROACH---------------------------------------------------

\section{Approach} \label{sec:approach}

We want to build upon the method of Robnik-{\v{S}}ikonja and Kononenko \yrcite{robnik2008explaining} to develop a tool for analyzing how DCNNs classify images. For a given input image, the method will allow us to estimate the importance of each pixel by assigning it a relevance value. The result can then be represented in an image which is of the same size as the input image. 

Our main contributions are three substantial improvements of this method: \emph{conditional sampling} (section \ref{sec:approach:conditional}), \emph{multivariate analysis} (section \ref{sec:approach:multivariate_analysis}), and \emph{deep visualization} (section \ref{sec:approach:deepvis}).

\subsection{Conditional Sampling} \label{sec:approach:conditional}

To adapt the method for the use with DCNNs, we will utilize the fact that in natural images, the value of a pixel does not depend so much on its coordinates, but much more on the pixels around it. The probability of a red pixel suddenly appearing in a clear-blue sky is rather low, which makes $p(x_i)$ not a very accurate approximation of $p(x_i|\xbacki)$. A much better approximation can be found by considering the following two observations: a pixel's value depends most strongly on the pixels in some neighborhood around it (and not so much on the pixels far away), and a pixel's value does not depend on the location of it in the image (in terms of coordinates).
For a pixel $x_i$, we can therefore find a patch $\hat{\mathbf{x}}_i$ of size $l \times l$ that contains $x_i$, and condition on the remaining pixels in that patch:
\begin{equation}
p(x_i|\xbacki) \approx p(x_i|\hat{\mathbf{x}}_{\backslash i}) \ .
\end{equation}
This greatly improves the approximation while remaining completely tractable.

For a feature to become relevant when using conditional sampling, it now has to satisfy two conditions: being relevant to predict the class of interest, and be hard to predict from the neighboring pixels. Relative to the marginal method, we therefore downweight the pixels that can easily be predicted and are thus redundant in this sense. 

\subsection{Multivariate Analysis} \label{sec:approach:multivariate_analysis}

Robnik-{\v{S}}ikonja and Kononenko \yrcite{robnik2008explaining} take a \textit{univariate} approach: only one feature at a time is removed. However, we would expect that a neural network is robust to just one feature of a high-dimensional input being unknown, like a pixel in an image. Ideally, we would marginalize out every element of the power set of features, but this is clearly unfeasible for high-dimensional data. Therefore, we will remove several features at once by again making use of our knowledge about images by strategically choosing these feature sets: patches of connected pixels. Instead of going through all individual pixels, we go through all patches of size $k\times k$ in the image, implemented in a sliding windows fashion. The patches are overlapping, so that ultimately an individual pixel's relevance is obtained by taking the average relevance obtained from the different patches it was in.

Algorithm \ref{alg:pred-diff} shows how the method can be implemented, incorporating the proposed improvements.

% --- ALGORITHM ---
\begin{algorithm}[t]
   \caption{Evaluating the Prediction Difference}
   \label{alg:pred-diff}
\begin{algorithmic}
   \INPUT a classifier function, input image $\mathbf{x}$ of size $n \times n$, inner patch size $k$, outer patch size $l>k$, class of interest $c$, probabilistic model over patches of size $l \times l$, number of samples $S$
   \STATE \textbf{initialization:} WE = zeros(n*n), counts = zeros(n*n)
   \FOR{ every patch $\mathbf{x}_w$ of size $k\times k$ {\bfseries in} $\mathbf{x}$}
     \STATE $\mathbf{x}' = \text{copy}(\mathbf{x})$
     \STATE $\text{sum}_w = 0$
   	 \STATE define patch $\hat{\mathbf{x}}_{w}$ of size $l\times l$ that contains $\mathbf{x}_w$
     \FOR{$s=1$ {\bfseries to} $S$}
       \STATE $\mathbf{x}'_w \leftarrow\mathbf{x}_w$ sampled from $p(\mathbf{x}_w|\hat{\mathbf{x}}_{w}\backslash \mathbf{x}_w)$
       \STATE evaluate the classifier to get $p(c|\mathbf{x}')$
       \STATE $\text{sum}_w \mathrel{+}= p(c|\mathbf{x}')$ 
     \ENDFOR
     \STATE $p(c|\mathbf{x}\backslash \mathbf{x}_w) := \text{sum}_w / S$
   	 \STATE WE[coordinates of $\mathbf{x}_w$] \\ \hspace{1.3cm} $ \mathrel{+}= \log_2(\text{odds}(c|\mathbf{x})) - \log_2(\text{odds}(c|\mathbf{x}\backslash \mathbf{x}_w))$ 
     \STATE counts[coordinates of $\mathbf{x}_w$] $\mathrel{+}= 1$
   \ENDFOR
   \OUTPUT WE / counts (point-wise)
\end{algorithmic}
\end{algorithm}
% -----------------

\subsection{Deep Visualization of Hidden Layers} \label{sec:approach:deepvis}

When trying to understand neural networks and how they make decisions, it is not only interesting to analyze the input-output relation of the classifier, but also to look at what is going on inside the hidden layers of the network. We can adapt the method to see how the units of any layer of the network influence a node from a deeper layer. 
Mathematically, we can formulate this as follows. Let $\mathbf{h}$ be the vector representation of the values in a layer $H$ in the network (after forward-propagating the input up to this layer).
Further, let $z = z(\mathbf{h})$ be the value of a node that depends on $\mathbf{h}$.
Then the analog of eq. (\ref{eq:marginal-exact}) is given by the expectation:
\begin{equation}
	g(z|\mathbf{h}_{\backslash i})
	\equiv \mathbb{E}_{p(h_i|\mathbf{h}_{\backslash i})} \left[ z(\mathbf{h}) \right] 
	= \sum\limits_{h_i} p(h_i|\mathbf{h}_{\backslash i} ) z( \mathbf{h}_{\backslash i}, h_i ) \ ,
\end{equation}
which expresses the distribution of $z$ when unit $h_i$ in layer $H$ is unobserved. 
The equation now works for arbitrary layer/unit combinations, and evaluates to the same as equation (\ref{eq:marginal-exact}) when the input-output relation is analyzed.
We still have to define how to evaluate the difference between $g(z|\mathbf{h})$ and $g(z|\mathbf{h}_{\backslash i})$. Since we are not necessarily dealing with probabilities anymore, equation (\ref{eq:weight-of-evidence}) is not suitable. In the general case, we will therefore just look at the \textit{activation difference},
\begin{equation} \label{eq:inst-activDiff}
\text{AD}_i(z|\mathbf{h}) = g(z|\mathbf{h}) - g(z|\mathbf{h}_{\backslash i}) \ .
\end{equation}

% ------------------- EXPERIMENTS ----------------------------------------------------------

\section{Experiments} \label{sec:experiments}

In this section, we illustrate how the proposed visualization method can be applied. We use images from the ILSVRC challenge \cite{ILSVRC15} (a large dataset of natural images from  $1000$ categories) and three DCNNs: the AlexNet \citep{krizhevsky2012imagenet}, the GoogLeNet \citep{szegedy2015going} and the (16-layer) VGG network \citep{simonyan2014very}. We used the publicly available pre-trained models that were implemented using the deep learning framework caffe \citep{jia2014caffe}.
Analyzing one image took us 0.5, 1 and 5 hours for the respective classifiers AlexNet, GoogLeNet and VGG (with $4$GB GPU memory and using mini-batches).

We compare our method to the sensitivity analysis by Simonyan et al. \yrcite{simonyan2013deep}. The sensitivity map with respect to a node (from the penultimate layer) is computed with a single backpropagation pass, handled by the caffe framework.

For marginal sampling we always use the empirical distribution, i.e., we replace a feature (patch) with samples taken directly from other images, at the same location. For conditional sampling we use a multivariate normal distribution. 
For each small pixel patch of size $k\times k\times 3$ (over all channels of the RGB image), we find a larger patch of size $l \times l \times 3$ that contains the smaller patch (and so that it is as centered as possible). Using a Gaussian model on the large patch, we can sample values for the small patch, conditioned on the remaining pixels. The model parameters (mean and covariance) were estimated using around $25,000$ patches of size $l \times l \times 3$ from the ImageNet data. 
For both sampling methods we use $20$ samples to estimate $p(c|\xbacki)$.

\subsection{Understanding how a DCNN makes decisions} \label{sec:experiments:explaining}

\begin{figure*}[t]
\includegraphics[width=0.8\textwidth]{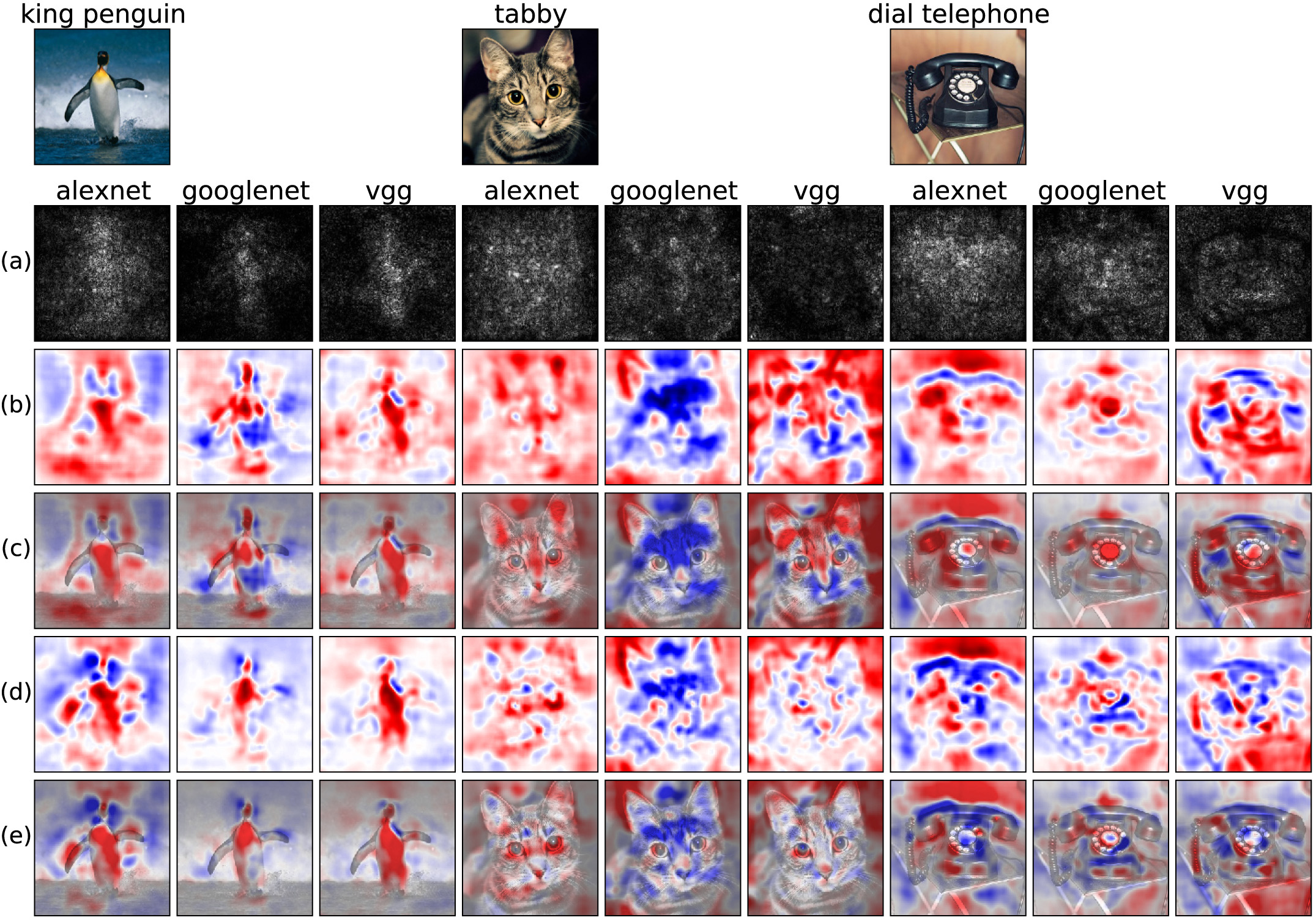}
\centering
\caption{\small \textbf{Visualization of the relevance of input features for the predicted class.} All networks make correct predictions, except AlexNet predicts tiger cat instead of tabby cat. Shown is the relevance of the input pixels on the highest predicted class. (a) shows the sensitivity map, (b) the prediction difference with marginal sampling, and (c) the same result overlaid with the input image. (d)+(e) show the results with conditional sampling. For each image, we show the results for the three networks AlexNet, GoogLeNet and VGG net (columns), using patch sizes $k=10$ and $l=14$ (see alg. \ref{alg:pred-diff}). The \textbf{colors} in the visualizations have the following meaning: \textit{red} stands for evidence for the predicted class; \textit{blue} regions are evidence against it. Transparent regions do not have an influence on the decision.} 
\label{fig:experiments:overview}
\end{figure*}

\begin{figure*}[t]
\includegraphics[width=0.8\textwidth]{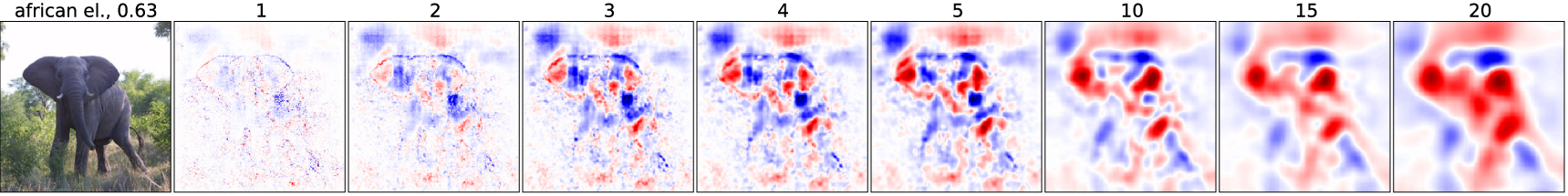}
\centering
\caption{\small \textbf{Visualization of how different patch sizes influence the result}, i.e., when $k$ in alg. \ref{alg:pred-diff} varies. We used the conditional sampling method and the AlexNet classifier, and set $l=k+4$ in all examples.}
\label{fig:experiments:patchsize}
\end{figure*}

Figure \ref{fig:experiments:overview} shows visualizations of the spatial support for the highest scoring class, for the three different classifiers, using marginal and conditional sampling.
Red areas indicate evidence \textit{for} the class, while blue indicates evidence \textit{against} the class. Transparent areas indicate that the pixels did not have any influence on the prediction.
For example, large parts of the cat's face are blue for the GoogLeNet, while the ear is red with high intensity.
This indicates that the classifier does not recognize the face as being indicative of the tabby cat class (but e.g. looks more like another cat class), while the ear appears very distinctive.

One obvious difference to the sensitivity map is that with our method, we have \textit{signed} information about the feature's relevance. (The partial derivatives are of course signed, but this encodes a different kind of information.) We can see that often, the sensitivity analysis highlights the class object in the image. Our method does not necessarily highlight the object itself, but the things that the classifier uses to detect what is in the image, which can also be contextual information.

When comparing marginal and conditional sampling, we see that in general, conditional sampling gives sharper results.
For the rest of our experiments, we will use conditional sampling only.

Comparing the visualizations of the three classifiers, we see that the explanations for their decisions differ. For example, we can see that in (d) for the penguin, the VGG network considers the penguin's head as evidence for the class, whereat the AlexNet considers it evidence against the class.

\subsection{Patch Size}

\begin{figure*}[t]
\includegraphics[width=0.9\textwidth]{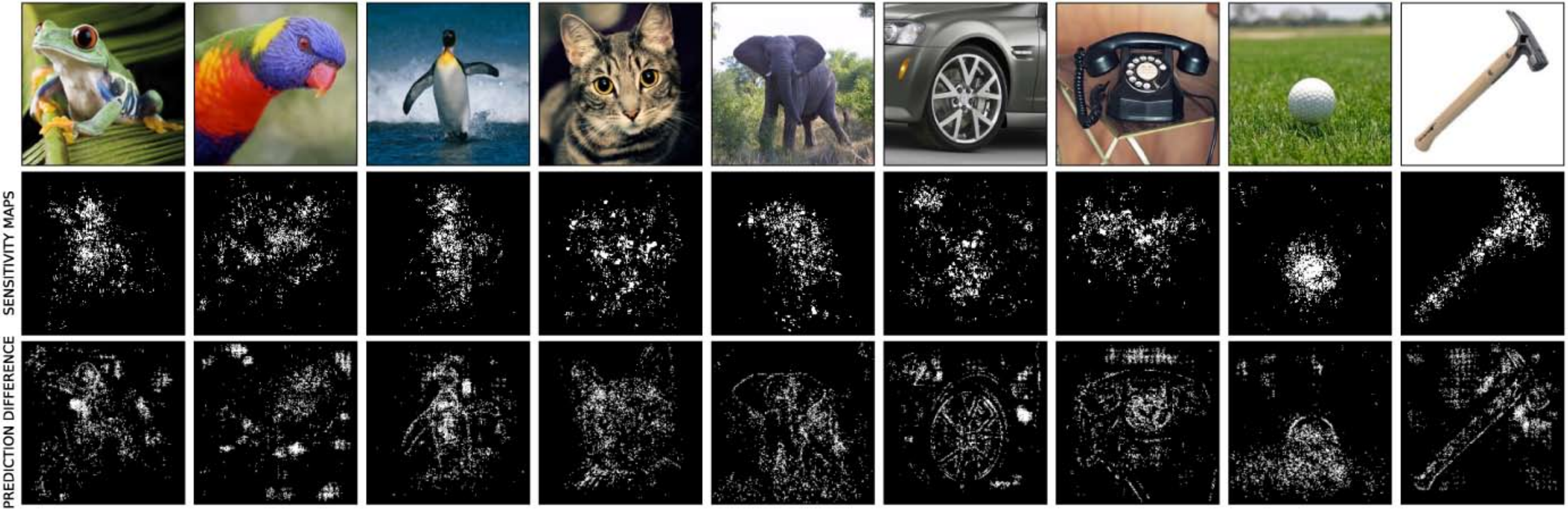}
\centering
\caption{\small \textbf{Top 5\% of the relevant pixels} by absolute value visualized by a boolean mask. The first row shows the results from the sensitivity map; the second row the results using our methods with conditional sampling on the AlexNet, using $k=1$ and $l=5$ in alg. 1 (i.e., marginalizing single pixels). Both methods show the visualization with respect to the highest scoring class in the penultimate layer.}
\label{fig:experiments:patchsize1}
\end{figure*}

\begin{figure*}[t]
\includegraphics[width=0.95\textwidth]{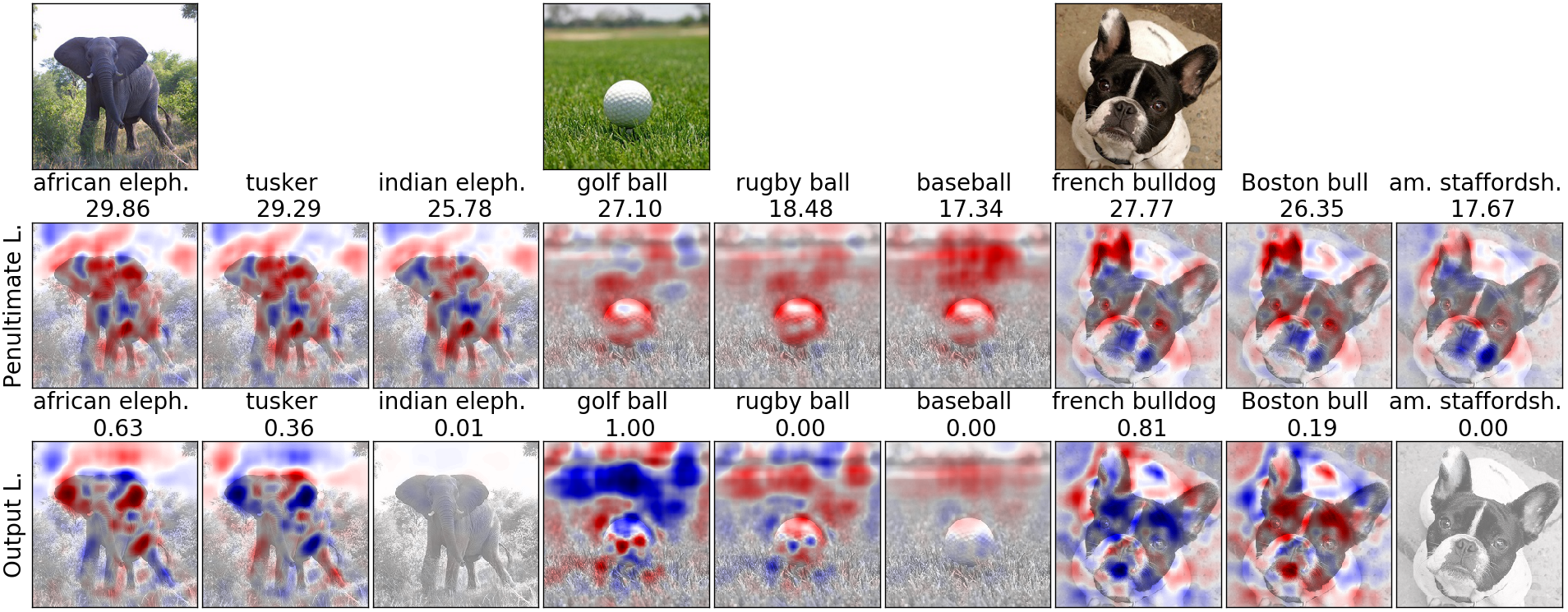}
\centering
\caption{\small \textbf{Visualization of the support for the top-three scoring classes in the penultimate layer and output layer}. The second row shows the results with respect to the penultimate layer; the third row with respect to the output layer. For each image, we additionally report the values of the units. We used the AlexNet with conditional sampling and patch sizes $k=10$ and $l=14$ (see alg. \ref{alg:pred-diff}). \textit{Red} pixels are evidence for a class, \textit{blue} against it.}
\label{fig:experiments:pen-vs-out}
\end{figure*}

In the previous examples, we used a patch size of $10\times 10\times 3$ for the set of pixels we marginalize out at once. In general, this gives a good trade-off between sharp results and a smooth appearance. But let us now look at different resolutions by varying the patch size for the image of the elephant in figure \ref{fig:experiments:patchsize} (note that these results are not just simple averages of one another, but a multivariate approach is indeed necessary to observe the presented results). Surprisingly, removing only one pixel (all three RGB values) has a effect on the prediction, and the largest effect comes from sensitive pixels. We expected that removing only one pixel does not have any effect on the classification outcome, but apparently the classifier is sensitive even to these small changes. When using such a small patch size, it is difficult to make sense of the sign information in the visualization. If we want to get a good impression of which parts in the image are evidence for/against a class, it is therefore better to use larger patches. 

The smaller patches have the advantage that the outlines of the object are highlighted very clearly. In figure \ref{fig:experiments:patchsize1}, we show the top $5\%$ relevant pixels by absolute value (in a boolean mask), using the sensitivity map and the prediction difference with $k=1$. The sensitivity map seems to mainly concentrate on the object itself, whereas our method regards the outlines of the object as most relevant.

\subsection{Pre-Softmax versus Output Layer}

If we visualize the influence of the input features on the penultimate (pre-softmax) layer, we show only the evidence for/against this particular class, without taking other classes into consideration. After the softmax operation however, the values of the nodes are all interdependent: a drop in the probability for one class could be due to less evidence for it, or because a different class becomes more likely.
Figure \ref{fig:experiments:pen-vs-out} compares visualizations for the last two layers. By looking at the top three scoring classes, we can see that the visualizations in the penultimate layer look very similar if the classes are similar (like different dog breeds).
When looking at the output layer however, they look rather different. Consider the case of the elephants: the top three classes are different elephant subspecies, and the visualizations of the penultimate layer look similar since every subspecies can be identified by similar characteristics. But in the output layer, we can see how the classifier decides for one of the three types of elephants and against the others: the ears in this case are the crucial difference.

\subsection{Deep Visualization of Hidden Network Layers}

\begin{figure*}[t]
\includegraphics[width=0.99\textwidth]{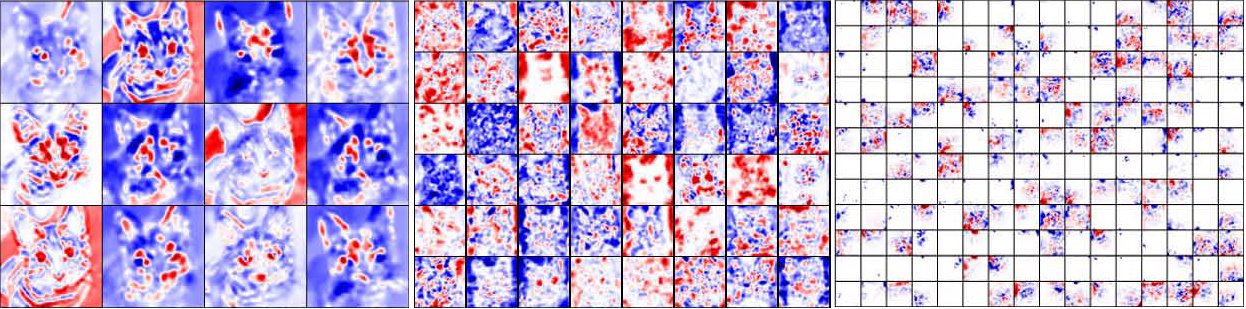}
\centering
\caption{\small Visualization of feature maps from thee different layers of the GoogLeNet (l.t.r.: ''conv1/7x7\_s2'', ''inception\_3a/output'', ''inception\_5b/output''), using conditional sampling and patch sizes $k=10$ and $l=14$ (see alg. \ref{alg:pred-diff}). For each feature map in the convolutional layer, we first evaluate the relevance for every single unit, and then average the results over all the units in one feature map to get a sense of what the unit is doing as a whole. \textit{Red} pixels activate a unit, \textit{blue} pixels decreased the activation.}
\label{fig:experiments:deepvis-1}
\end{figure*}

\begin{figure*}[t]
\includegraphics[width=0.78\textwidth]{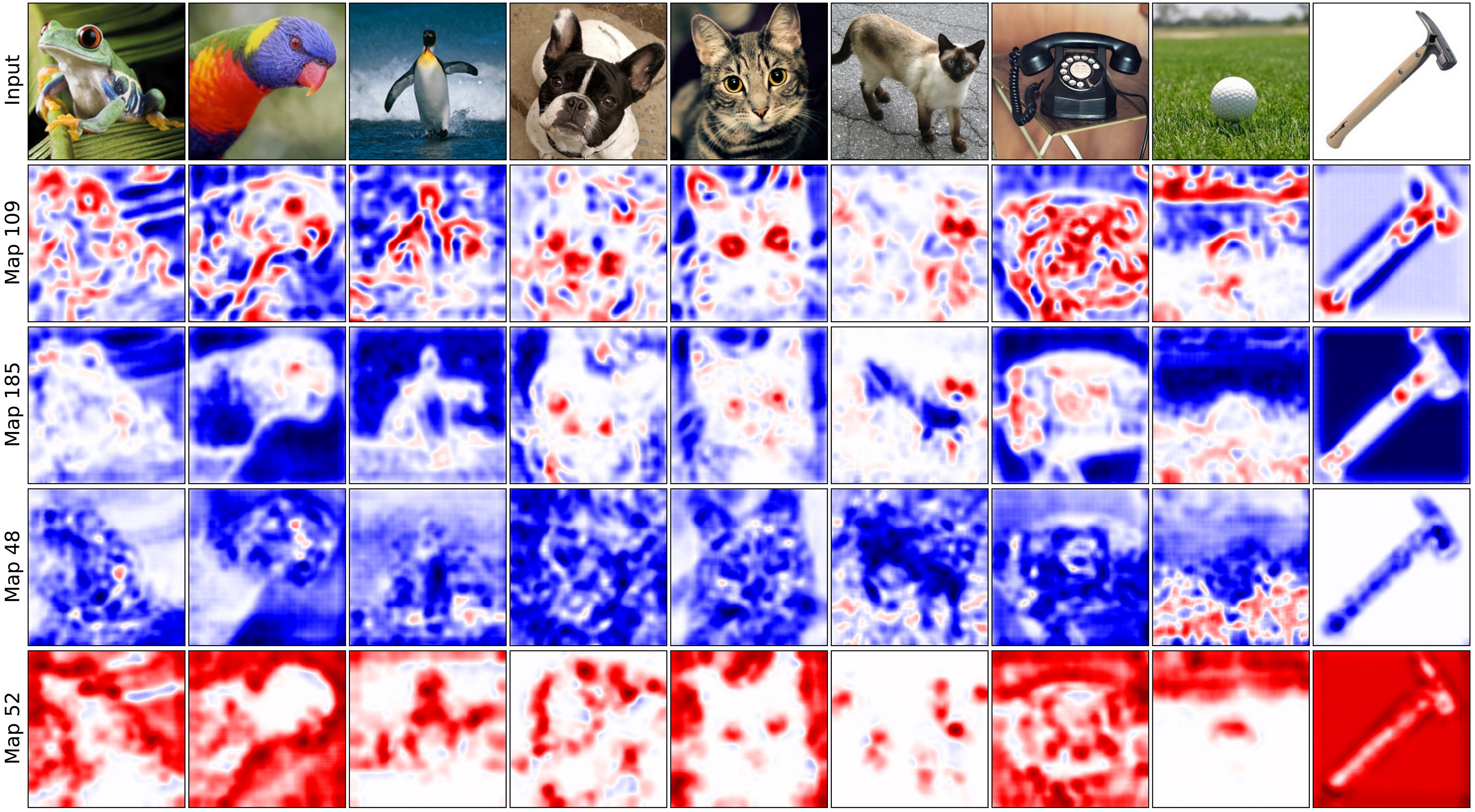}
\centering
\caption{\small \textbf{Visualization of four different feature maps}, taken from the ''inception\_3a/output'' layer of the GoogLeNet (from the middle of the network). Shown is the average relevance of the input features over all activations of the feature map. We used patch sizes $k=10$ and $l=14$ (see alg. \ref{alg:pred-diff}). \textit{Red} pixels activate a unit, \textit{blue} pixels decreased the activation.}
\label{fig:experiments:deepvis-2}
\end{figure*}

Section \ref{sec:approach:deepvis} illustrated how our method can be used to understand the role of hidden layers of a DNN. Figure \ref{fig:experiments:deepvis-1} shows how different feature maps in three different layers of the GoogLeNet react to the input of the tabby cat from figure \ref{fig:experiments:overview}. For each feature map in a convolutional layer, we first compute the relevance of the input image for each hidden unit in the map, and then average over the results for all the units to visualize what the feature map as a whole is doing. We can thus see which parts of the input image respond positively or negatively to these feature maps. 
The first convolutional layer works with different types of simple image filters (e.g., edge detectors), and what we see is which parts of the input image respond positively or negatively to these filters. The layer we picked from somewhere in the middle of the network is specialized to higher level features (like facial features of the cat).
The activations of the last convolutional layer are very sparse across feature channels, indicating that these units are highly specialized.

To get a sense of what single feature maps in convolutional layers are doing, we can look at their visualization for different input images and search for patterns in their behavior. 
Figure \ref{fig:experiments:deepvis-2} shows this for four different feature maps from a layer from the middle of the GoogLeNet network. Here, we can directly see which kind of features the model has learned at this stage in the network. For example, one feature map is is activated by the eyes of animals (second row), and another is looking mostly at the background (last row).

% ------------------- CONCLUSION ------------------------------------------------------------

\section{Conclusion}

We have presented a new method for visualizing deep neural networks that improves on previous methods by using a more powerful conditional, multivariate model. The visualization method shows which pixels of a specific input image are evidence for or against a node in the network. Compared to the sensitivity analysis, the signed information offers new insights - for research on the networks, as well as the acceptance and usability in domains like healthcare. In our experiments, we have presented several ways in which the visualization method can be put into use for analyzing how DCNNs make decisions. For future work, we can imagine that using more sophisticated generative models (instead of a simple multivariate normal distribution) will lead to better results: pixels that are easily predictable by their surrounding are downweighted even more.

\section*{Acknowledgments} 
This work was supported by AWS in Education Grant award. We also thank Facebook and Google.

\bibliography{conv_vis}

\begin{thebibliography}{11}
\providecommand{\natexlab}[1]{#1}
\providecommand{\url}[1]{\texttt{#1}}
\expandafter\ifx\csname urlstyle\endcsname\relax
  \providecommand{\doi}[1]{doi: #1}\else
  \providecommand{\doi}{doi: \begingroup \urlstyle{rm}\Url}\fi

\bibitem[Erhan et~al.(2009)Erhan, Bengio, Courville, and
  Vincent]{erhan2009visualizing}
Erhan, Dumitru, Bengio, Yoshua, Courville, Aaron, and Vincent, Pascal.
\newblock Visualizing higher-layer features of a deep network.
\newblock \emph{Dept. IRO, Universit{\'e} de Montr{\'e}al, Tech. Rep}, 4323,
  2009.

\bibitem[Jia et~al.(2014)Jia, Shelhamer, Donahue, Karayev, Long, Girshick,
  Guadarrama, and Darrell]{jia2014caffe}
Jia, Yangqing, Shelhamer, Evan, Donahue, Jeff, Karayev, Sergey, Long, Jonathan,
  Girshick, Ross, Guadarrama, Sergio, and Darrell, Trevor.
\newblock Caffe: Convolutional architecture for fast feature embedding.
\newblock \emph{arXiv preprint arXiv:1408.5093}, 2014.

\bibitem[Krizhevsky et~al.(2012)Krizhevsky, Sutskever, and
  Hinton]{krizhevsky2012imagenet}
Krizhevsky, Alex, Sutskever, Ilya, and Hinton, Geoffrey~E.
\newblock Imagenet classification with deep convolutional neural networks.
\newblock In \emph{Advances in neural information processing systems}, pp.\
  1097--1105, 2012.

\bibitem[Robnik-{\v{S}}ikonja \& Kononenko(2008)Robnik-{\v{S}}ikonja and
  Kononenko]{robnik2008explaining}
Robnik-{\v{S}}ikonja, Marko and Kononenko, Igor.
\newblock Explaining classifications for individual instances.
\newblock \emph{Knowledge and Data Engineering, IEEE Transactions on},
  20\penalty0 (5):\penalty0 589--600, 2008.

\bibitem[Russakovsky et~al.(2015)Russakovsky, Deng, Su, Krause, Satheesh, Ma,
  Huang, Karpathy, Khosla, Bernstein, Berg, and Fei-Fei]{ILSVRC15}
Russakovsky, Olga, Deng, Jia, Su, Hao, Krause, Jonathan, Satheesh, Sanjeev, Ma,
  Sean, Huang, Zhiheng, Karpathy, Andrej, Khosla, Aditya, Bernstein, Michael,
  Berg, Alexander~C., and Fei-Fei, Li.
\newblock {ImageNet Large Scale Visual Recognition Challenge}.
\newblock \emph{International Journal of Computer Vision (IJCV)}, 115\penalty0
  (3):\penalty0 211--252, 2015.
\newblock \doi{10.1007/s11263-015-0816-y}.

\bibitem[Simonyan \& Zisserman(2014)Simonyan and Zisserman]{simonyan2014very}
Simonyan, Karen and Zisserman, Andrew.
\newblock Very deep convolutional networks for large-scale image recognition.
\newblock \emph{arXiv preprint arXiv:1409.1556}, 2014.

\bibitem[Simonyan et~al.(2013)Simonyan, Vedaldi, and
  Zisserman]{simonyan2013deep}
Simonyan, Karen, Vedaldi, Andrea, and Zisserman, Andrew.
\newblock Deep inside convolutional networks: Visualising image classification
  models and saliency maps.
\newblock \emph{arXiv preprint arXiv:1312.6034}, 2013.

\bibitem[Szegedy et~al.(2015)Szegedy, Liu, Jia, Sermanet, Reed, Anguelov,
  Erhan, Vanhoucke, and Rabinovich]{szegedy2015going}
Szegedy, Christian, Liu, Wei, Jia, Yangqing, Sermanet, Pierre, Reed, Scott,
  Anguelov, Dragomir, Erhan, Dumitru, Vanhoucke, Vincent, and Rabinovich,
  Andrew.
\newblock Going deeper with convolutions.
\newblock In \emph{Proceedings of the IEEE Conference on Computer Vision and
  Pattern Recognition}, pp.\  1--9, 2015.

\bibitem[Yosinski et~al.(2015)Yosinski, Clune, Nguyen, Fuchs, and
  Lipson]{yosinski2015understanding}
Yosinski, Jason, Clune, Jeff, Nguyen, Anh, Fuchs, Thomas, and Lipson, Hod.
\newblock Understanding neural networks through deep visualization.
\newblock \emph{arXiv preprint arXiv:1506.06579}, 2015.

\bibitem[Zeiler \& Fergus(2014)Zeiler and Fergus]{zeiler2014visualizing}
Zeiler, Matthew~D and Fergus, Rob.
\newblock Visualizing and understanding convolutional networks.
\newblock In \emph{Computer vision--ECCV 2014}, pp.\  818--833. Springer, 2014.

\bibitem[Zhou et~al.(2014)Zhou, Khosla, Lapedriza, Oliva, and
  Torralba]{zhou2014object}
Zhou, Bolei, Khosla, Aditya, Lapedriza, Agata, Oliva, Aude, and Torralba,
  Antonio.
\newblock Object detectors emerge in deep scene cnns.
\newblock \emph{arXiv preprint arXiv:1412.6856}, 2014.

\end{thebibliography}
\bibliographystyle{icml2016}
\end{document}